\documentclass{edm_article}
\usepackage{xcolor}

\begin{document}
\title{qDKT: Question-centric Deep Knowledge Tracing}
\numberofauthors{5}
\author{
\alignauthor
        Shashank Sonkar\\
        \affaddr{Rice University}\\
        \email{ss164@rice.edu}
\alignauthor
        Andrew E. Waters\\
        \affaddr{OpenStax/ Rice University}\\
        \email{aew2@rice.edu}
\and
\alignauthor
        Andrew S. Lan\\
        \affaddr{University of Massachusetts Amherst}\\
        \email{andrewlan@cs.umass.edu }
\alignauthor
        Phillip J. Grimaldi \\
        \affaddr{OpenStax/ Rice University}\\
        \email{pjg3@rice.edu}
\alignauthor
Richard G. Baraniuk\\
        \affaddr{Rice University}\\
        \email{richb@rice.edu}
}
\maketitle
\begin{abstract}

Knowledge tracing (KT) models, e.g., the deep knowledge tracing (DKT) model, track an individual learner's acquisition of skills over time by examining the learner's performance on questions related to those skills. A practical limitation in most existing KT models is that all questions nested under a particular skill are treated as equivalent observations of a learner's ability, which is an inaccurate assumption in real-world educational scenarios. To overcome this limitation we introduce qDKT, a variant of DKT that models every learner's success probability on individual questions over time. First, qDKT incorporates graph Laplacian regularization to smooth predictions under each skill, which is particularly useful when the number of questions in the dataset is big. Second, qDKT uses an initialization scheme inspired by the fastText algorithm, which has found success in a variety of language modeling tasks. Our experiments on several real-world datasets show that qDKT achieves state-of-art performance on predicting learner outcomes. Because of this, qDKT can serve as a simple, yet tough-to-beat, baseline for new question-centric KT models.
\end{abstract}

\section{Introduction}
Knowledge tracing (KT) models are useful tools which provide educators with actionable insights into learners' progress \cite{vanlehn2006behavior, pelanek2017bayesian}. Given a learner's performance history, these methods predict their proficiency across a predetermined set of skills (i.e., knowledge components or concepts). One of the most popular methods for tracking this cognitive development is the Bayesian Knowledge Tracing (BKT) framework \cite{corbett1994knowledge, pardos2010navigating, yudelson2013individualized} which applies hidden Markov models \cite{baum1966statistical} to learn each learner's \textit{guess, slip,} and \textit{learn} probabilities for each skill. Another approach to modeling the dynamics of skill acquisition is SPARFA-Trace \cite{lan2014time} which uses Kalman filtering \cite{kalman1960new} to model learner skill acquisition. An advantage of SPARFA-Trace is that it can, unlike BKT models, relate individual questions to multiple skills. Recently, deep learning techniques have been applied to the KT problem to create Deep Knowledge Tracking (DKT) \cite{piech2015deep} which models the sequence prediction task using a Long Short-Term Memory (LSTM) network \cite{hochreiter1997long}.

All of the aforementioned KT models track an individual learner's knowledge at the \textit{skill} level. Under the KT framework, the time series data modeled consists of learner skill interaction sequences, given by $X_i = \{ (s_t^i, a_t^i) \}_{t=1}^{T}$ where $s_t^i$ is the skill index attempted by the $i^{\text{th}}$ learner at discrete time step $t$, while $a_t^i \in \{0,1\}$ is the assessment of the learner's response, with $0$ indicating an incorrect response and $1$ indicating a correct response. 

The key assumption underpinning these models is that all questions nested under a particular skill are equivalent. This assumption, however, is generally unrealistic in real-world educational datasets. First, a mapping of questions to skills is not always available and obtaining such a mapping requires the intervention of subject matter experts, which is both costly and time-consuming.  Second, questions in  real-world educational datasets are never homogeneous, but rather exhibit significant variations in difficulty and discrimination \cite{embretson2013item}.  In other words, different questions convey differing levels of information about a particular learner's mastery of the underlying skill, and methods for modeling learner's acquisition of skills over time should take such information into account. 

However, simply substituting questions for skills in a traditional KT model is insufficient to accomplish this goal. To illustrate this, we selected two commonly used educational datasets,  ASSISTments2009 and ASSISTments2017\footnote{https://sites.google.com/site/assistmentsdata/home}. We first ran the standard DKT model using the skill-level information provided with each dataset.  We then re-ran the DKT model but used the question identifiers themselves, rather than the skills, for modeling performance. Concretely, the time series data modeled consisted of learners' question interaction sequences, given by $X_i = \{ (q_t^i, a_t^i) \}_{t=1}^{T}$, where $q_t^i$ denotes the question answered by learner $i$ at time $t$. The AUC for both of these model variants are shown in Table~\ref{tab:tab1}.  We note that for the ASSISTments 2017 dataset that this question-centric approach provides a moderate improvement in AUC but for the ASSISTments 2009 dataset the question-centric approach significantly hurt AUC. To understand why this behavior occurs we note that the average number of observations per question for the ASSISTments 2009 dataset is significantly smaller than that for the ASSISTments 2017 dataset. This results in the question-centric modeling overfitting to the data, which adversely affects predictive accuracy. In contrast, the ASSISTments 2017 dataset has a larger number of observations per question, which helps the question-centric DKT model to avoid overfitting.

\begin{table}[ht]
\centering
\resizebox{1\columnwidth}{!}{
\begin{tabular}{|c|c|c|c|c|}
\hline
\textbf{Dataset} &
\textbf{\begin{tabular}[c]{@{}c@{}}Number of\\questions\end{tabular}} &
\textbf{\begin{tabular}[c]{@{}c@{}}Avg. Obs. \\per question\end{tabular}} &
\textbf{\begin{tabular}[c]{@{}c@{}}DKT\\(skill)\end{tabular}} & \textbf{\begin{tabular}[c]{@{}c@{}}DKT\\(question)\end{tabular}} \\ \hline
ASSISTments 2017 & 1,183  & 145.76             & 0.72         & \textbf{0.74}          \\ \hline
ASSISTments 2009 & 16,891 & 19.27            & \textbf{0.74}        & 0.68          \\ \hline
\end{tabular}
}
\caption{AUC scores for DKT vs its variant with questions as indices. Using questions indices leads to overfitting when the number of observations per question is small.}
\label{tab:tab1}
\end{table}

It is apparent that question-level modeling has the potential for significant improvement in predictive accuracy over skill-level modeling in KT models; however, simply substituting questions for skills in KT models is insufficient to achieve this. Addressing this challenge is the focus of our work and our main contributions are summarized as follows:

\begin{enumerate}
    \item We propose a novel algorithm for question-level know\-ledge tracing, which we dub qDKT, that achieves state-of-the-art performance compared to traditional KT \\ methods on a number of real-world datasets.
    \item Our method utilizes a novel graph Laplacian regularizer for incorporating question similarity information into qDKT. Question similarity can be calculated using the skill information or using textual similarity measures if the dataset contains the actual text for each question. Unlike other KT methods, our method does not assume that each question must be associated with exactly one skill.
    \item We propose a novel initialization scheme for question-level KT models using fastText \cite{bojanowski2017enriching}, an algorithm for natural language processing (NLP). This initialization scheme learns embeddings that summarize pointwise mutual information statistics\cite{levy2014neural}, which is beneficial for bootstrapping sequence prediction models.
\end{enumerate}

Incorporating question-information to improve skill-centric KT models have been tried in the past, for example, the model proposed by \cite{wangdeep} concatenates the question embedding to the skill embedding, which is then used as the input to the model. As training progresses, the model learns both the question embedding, and the skill embedding. However, the focus of our proposed initialization scheme is to bootstrap question-centric KT models without using any skill information. As stated earlier, this is advantageous because firstly, tagging questions with skills can be expensive, and secondly, the design of current skill-centric KT models does not transfer well to question-centric KT models (as shown in table~\ref{tab:tab1}). Initialized with the fastText-inspired scheme, qDKT performs at par with the state-of-art skill-level DKT model on ASSISTments 2009 dataset, and improves it by 5\% and 6\% on ASSISTments 2017 dataset and Statics 2011 dataset respectively.

Coupling the fastText-inspired scheme with the Laplacian regularizer, qDKT gives gains of 2\% in AUC score as compared to the  skill-centric DKT model for ASSISTments 2009, while also capturing question-specific characteristics.

\section{Problem Statement and DKT \\ overview}
Each learner's performance record contains the questions attempted, time at which each question was attempted, and the assessment of each response (either correct or incorrect). Also, assume that the skill associated with every question is known. Given performance records for several learners, one wishes to train a knowledge tracing model with the objective of predicting the success probabilities across the questions (or the skills) at time $T$ for a new learner whose performance history has been recorded until time $T-1$.

\subsection{DKT Model}
DKT uses an LSTM to predict a learner's future performance using their previous assessment history. As discussed earlier, the input to the model is a time series which consists of learners' skill interaction sequences, given by $X_i = \{ (s_t^i, a_t^i) \}_{t=1}^{T}$. Here we restrict our discussion to a single learner and will omit the superscript $i$ throughout. The forward equations of the DKT model are given in \eqref{eq:1} -- \eqref{eq:before_loss}: 
\begin{equation} \label{eq:1}
\boldsymbol{x}_t = W_{xv}\boldsymbol{v}_t,
\end{equation}
\begin{equation}
\boldsymbol{i}_t = \sigma(W_i\boldsymbol{x}_t + U_i\boldsymbol{h}_{t-1} + \boldsymbol{b}_i),
\end{equation}
\begin{equation}
\boldsymbol{f}_t = \sigma(W_f\boldsymbol{x}_t + U_f\boldsymbol{h}_{t-1} + \boldsymbol{b}_f),
\end{equation}
\begin{equation}
\boldsymbol{o}_t = \sigma(W_o\boldsymbol{x}_t + U_o\boldsymbol{h}_{t-1} + \boldsymbol{b}_o),
\end{equation}
\begin{equation}
\boldsymbol{c}_t = \boldsymbol{f}_t \odot \boldsymbol{c}_{t-1} + \boldsymbol{i}_t \odot \sigma'(W_c\boldsymbol{x}_t + U_c\boldsymbol{h}_{t-1} + \boldsymbol{b}_c),
\end{equation}
\begin{equation}
\boldsymbol{h}_t = \boldsymbol{o}_t \odot \sigma'(\boldsymbol{c}_t),
\end{equation}
\begin{equation} \label{eq:before_loss}
\boldsymbol{y}_t = \sigma (W_{yh}\boldsymbol{h}_t + \boldsymbol{b}_y),
\end{equation}
where $\sigma$ is the sigmoid function, $\sigma'$ is the hyperbolic tangent function, and the operator $\odot$ denotes the element-wise multiplication.  In words, the input at time step $t$ is the skill interaction tuple $(s_t, a_t)$ which is encoded by an arbitrary high-dimensional one-hot vector, $\boldsymbol{v}_t \in \{0,1\}^{2M}$, where $M$ is the number of skills. Using an embedding matrix, $W_{xv} \in R^{K \times 2M}$, $\boldsymbol{v}_t$ is mapped to a low-dimensional vector, $\boldsymbol{x}_t \in R^K, K \ll M$ \eqref{eq:1}, which serves as the input to the LSTM cell. $\boldsymbol{x}_t$ is passed through each of the input, forget, and output gates and, in the end, the LSTM returns $\boldsymbol{h}_t$ -- the estimate of the learner's current knowledge state. The final output of the model is $\boldsymbol{y}_t \in R^M$ which predicts the learner's success probabilities for all the $M$ skills for the next time step $t+1$.

\subsubsection{Loss in the DKT Model.}
The output of the DKT model, $\boldsymbol{y}_t$, predicts the learner's proficiency over the skills for the next time step $t+1$. During training, the assessment ($a_{t+1}$) of the learner's response to the question indexed by $q_{t+1}$ is known beforehand. The success probability for the skill associated with $q_{t+1}$ is given by $y_t[s_{t+1}]$. Since DKT assumes that mastery in the skill is equivalent to mastery in any of the questions under it (i.e., all questions under a skill are equivalent), a trained DKT model should predict the success probability at the skill to be the same as the assessment. This rationale motivates the basis for calculating the loss, $\ell_t$, at time $t$, given by:
\begin{equation} \label{eq:loss}
\ell_t = l (y_t[s_{t+1}], a_{t+1}),
\end{equation}
where $\ell$ is binary cross-entropy loss.
\subsection{Proposed Model - qDKT}
\label{sec:qdkt}
We now introduce our proposed method for KT modeling at the question-level, which we dub qDKT. Our method considers a modified problem statement where we estimate a learner's success probability for each question rather than for each skill. Let a learner's question interaction sequence $X = \{(q_t, a_t) \}_{t=1}^{T-1}$ until time step $T-1$ be given, where $q_t$ denotes the question answered at time $t$ and $a_t \in \{0,1\}$ is the assessment of the response to question $q_t$. Our goal is to output $\boldsymbol{y}_t \in R^N$ which predicts the learner's success probabilities for all the $N$ questions at the next time step $t+1$. qDKT utilizes the same architecture as DKT as specified in \eqref{eq:1} - \eqref{eq:before_loss}, but with $\boldsymbol{v}_t \in \{0,1\}^{2N}$, $W_{xv} \in R^{K \times 2N}$, and $\boldsymbol{y} \in R^N$. The updated loss $\ell_t$ from \eqref{eq:loss} at time $t$ is then given by:
\begin{equation} \label{eq:qdkt_loss}
\ell_t = l (y_t[q_{t+1}], a_{t+1}).
\end{equation}

We shall refer to this model as the base qDKT model where the prefix $q$ represents question-level modeling.

\section{Regularization for \lowercase{q}DKT}
As seen in Table~\ref{tab:tab1}, the base qDKT model performs poorly for datasets with both a large number of questions and a small number of observations per question. To overcome this, we propose a regularization method for qDKT to combat overfitting. It is reasonable to assume that success probabilities of multiple questions associated with the same skill should not be significantly different for a given learner. Based on this premise, we regularize the variance in success probabilities for questions that fall under the same skill.
\begin{equation} \label{eq:reg}
    R(\boldsymbol{y})=  \sum_{i \in Q} \sum_{j \in Q} \boldsymbol{1}(i,j)\cdot(y_i - y_j)^2,
\end{equation}
where  vector $\boldsymbol{y} \in R^N$ contains success probabilities of all questions $Q$ in the dataset, $i, j \in Q$ and $\boldsymbol{1}(i,j)$ is 1 if $i, j$ fall under the same skill, otherwise it is 0.

We add this penalty to the loss and use $\lambda$ to control the weight of the penalty. Thus, the updated loss function from \eqref{eq:loss} with the regularization penalty is:
\begin{equation} \label{eq:lap_loss}
\ell = l + \lambda \cdot R (\boldsymbol{y}).
\end{equation}

\subsubsection{Interpretation of the regularizer.} Graph theory provides a clean interpretation for the regularization penalty which is also helpful for speeding up its computation. We construct a graph $G$ with number of nodes equal to the number of questions in the dataset. Two nodes are connected with an edge of weight $1$ if the questions are associated with the same skill and with an edge weight of $0$ otherwise. 

The degree matrix $D$ of a graph $G$ is a diagonal matrix with
\begin{equation*}
    d_{ii} = \sum_{j \in C_i} w_{ij},
\end{equation*}
where $w_{ij}$ is the similarity between node $i$ and node $j$ (edge weight), $C$ is the set containing all the indices directly connected with $i$ (immediate siblings). The adjacency matrix $A$ of a graph $G$ stores the edge weights $w_{ij}$.
Given the degree matrix $D$ and the adjacency matrix $A$ of a graph $G$, the Laplacian matrix $L$ is defined as:
\begin{equation*}
    L = D - A.
\end{equation*}
Then for any vector $\boldsymbol{v}$  \cite{hand2010statistical},
\begin{equation} \label{eq:simp}
    \boldsymbol{v}^T L \boldsymbol{v}=  \sum_{i, j} w_{ij} \cdot (v_i - v_j)^2.
\end{equation}
We can then use \eqref{eq:simp} to simplify the regularization penalty of \eqref{eq:reg}:
\begin{equation}
    R(\boldsymbol{y})=  \sum_{i \in Q} \sum_{j \in Q} \boldsymbol{1}(i,j)\cdot(y_i - y_j)^2 = \boldsymbol{y}^T L \boldsymbol{y} .
\end{equation}

The simplification of the double summation term to a condensed vector-matrix multiplication term is useful to speed up its calculation, especially while training the qDKT model on GPUs.

Further, our approach to model similarity works even when questions are associated with multiple skills. This provides additional flexibility over previous KT models that restrict each question to be associated to exactly one skill. Such flexibility is important for real-world applications where questions commonly evaluate learners on multiple skills simultaneously. Moreover, this formulation can be helpful to incorporate even other measures of similarity like tf-idf similarity \cite{martin2009speech} using question text.
\begin{table*}[t]
\centering
\resizebox{2.05\columnwidth}{!}{
\begin{tabular}{|c|c|c|c|c|c|c|}
\hline
\textbf{Dataset} & \textbf{\#Learners} & \textbf{\#Questions} & \textbf{\#Skills} & \textbf{\begin{tabular}[c]{@{}c@{}}Total learner\\ interaction tuples\end{tabular}} & \textbf{\begin{tabular}[c]{@{}c@{}}Unique skill\\ interaction tuples\end{tabular}} & \textbf{\begin{tabular}[c]{@{}c@{}}Unique question\\ interaction tuples\end{tabular}} \\ \hline
ASSISTments 2009  & 4,151 & 16,891 & 111 & 325,637 & 29,287 & 221\\ \hline
ASSISTments 2017  & 1,709 & 1,183 & 86 & 249,105 & 2,201 & 171 \\ \hline
Statics2011 & 333 & 1,223 & 85 & 189,297 & 190 & 2,446 \\ \hline
Tutor & 895 & 5,981 & 1,592 & 437,524 & 11,622 & 3,184 \\ \hline
\end{tabular}
}
\caption{Table summarizing dataset statistics.}
\label{tab:my-table}
\end{table*}

\section{Initialization for \lowercase{q}DKT}
\label{sec:intialization}
DKT maps each skill interaction tuple to $\boldsymbol{x} \in R^d$ via the matrix $W_{xv}$ (see \eqref{eq:1}). In DKT, the entries of $W_{xv}$ are initialized with draws from a standard normal distribution. While this approach is straightforward, random embeddings tend to perform extremely poorly in high dimensions where the optimization problem will have an extremely large number of saddle points \cite{dauphin2014identifying}. To overcome this limitation, we propose a more effective method for initializing $W_{xv}$ inspired by the fastText architecture.

\subsection{Overview of Language Modeling and fastText}
In NLP, language models are used to predict the most likely words that can follow a given sequence of words. Such models are often initialized with word embeddings from algorithms like word2vec \cite{mikolov2013distributed}, fastText and GloVe \cite{pennington2014glove}. At a high level, these algorithms embed words into a high dimensional space such that words that have close semantic relationships will be embedded near one another, while words with low semantic similarity will be embedded further apart \cite{goldberg2014word2vec}.

A novelty of fastText is that it considers individual characters in a word when computing the final embeddings. By doing this, fastText recognizes that the words ``love'', ``loved'', ``lovely'', and ``lovable'' are all related and embed them accordingly.

\subsection{Embedding Methods for Educational Response Data}

In our application, we wish to have a notion of question similarity that can serve to guide our initialization scheme, similar to the notion of similar word contexts in fastText.  To do this, we assemble an approximate ``text corpus'' from our response data, as follows:

Let set $Q$ contain all the question ids and set $U$ contain all characters. We define a one-to-one mapping $f : Q \to U$ which maps a question id to a unique character. To convert learners' question interaction sequences, $X = \{(q_t, a_t)\}_{t=1}^T$ into a text corpus, we apply a signal transformation $Y$ on $X$ such that $y_t = f(q_t) + a_t$ where `$+$' denotes the string concatenation operator. Thus, each question interaction is encoded as a two character string consisting of the question id and the graded response. This interaction encoding constitutes the ``words'' of our corpus. The ``sentences'' of our corpus constitute of the string of such encoded interactions by an individual learner. We finally apply fastText to this newly generated ``corpus''. For a given question interaction say $(q, 0)$, fastText will train the embeddings of the following $n$-grams $\{f(q), `0\textrm', f(q) + `0\textrm'\}$. Thus, we link the embeddings of $(q, 0)$ and $(q, 1)$ through the embedding of $f(q)$. The resulting output embedding of fastText is used as our initialization of $W_{xv}$.

Our rationale for this embedding scheme comes from the nature of educational data. Learner question interaction data tends to cluster into components of learners correctly and incorrectly answering certain subsets of questions based on their mastery of underlying subject material \cite{lan2014sparse}. By employing fastText, we create an embedding that is consistent with this feature of educational data to create a robust initialization that is more internally consistent with our data than a purely random initialization.

\begin{table*}[t]
\centering
\resizebox{2.05\columnwidth}{!}{
\begin{tabular}{|c|c|c|c|c|c|c|}
\hline
\textbf{Dataset} & \textbf{DKT} & \textbf{Base qDKT} & 
\textbf{\begin{tabular}[c]{@{}c@{}}Base qDKT w/\\ Laplacian regularizer\end{tabular}} &
\textbf{\begin{tabular}[c]{@{}c@{}}Base qDKT w/ \\ fastText \end{tabular}} &
 \textbf{\begin{tabular}[c]{@{}c@{}c@{}}Base qDKT w/\\ fastText \\ and regularizer\end{tabular}} \\ \hline
ASSISTments 2009 & 0.740 $\pm$ 0.002 & 0.678 $\pm$ 0.004 & 0.738 $\pm$ 0.003 & 0.740 $\pm$ 0.004  & \textbf{0.762 $\pm$ 0.005} \\ \hline
ASSISTments 2017 & 0.721 $\pm$ 0.002 & 0.742 $\pm$ 0.003 & 0.753 $\pm$ 0.005 & \textbf{0.772 $\pm$ 0.004}   & 0.770 $\pm$ 0.005 \\ \hline
Statics 2011 & 0.770 $\pm$ 0.003 & 0.822 $\pm$ 0.003 & 0.825 $\pm$ 0.002 & 0.832 $\pm$ 0.003 & \textbf{0.834 $\pm$ 0.002} \\ \hline
Tutor  & 0.856 $\pm$ 0.003 & 0.875 $\pm$ 0.002 & 0.882 $\pm$ 0.001 & 0.890 $\pm$  0.0008 & \textbf{0.895 $\pm$ 0.001}  \\ \hline
\end{tabular}
}
\caption{AUC scores for each algorithm and dataset. We see that both the addition of the regularizer and the improved initialization scheme improve performance on all datasets over the original DKT model. Combining both the regularizer and our proposed initialization scheme achieves the best performance over all algorithms.}

\label{tab:my-table2}
\end{table*}

\begin{figure*}[t]
\centering
\includegraphics[width=1\textwidth]{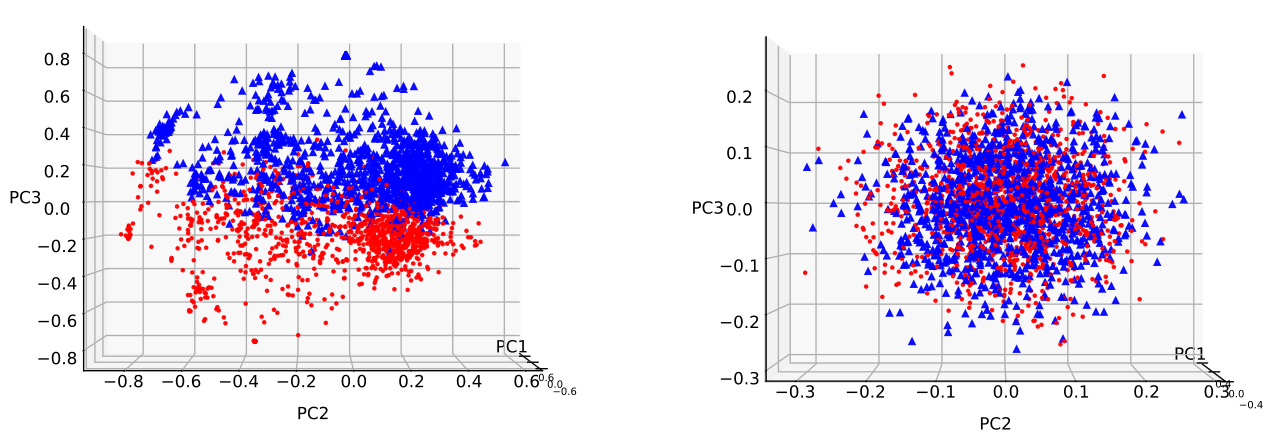}
\caption{PCA visualization of the columns of $W_{xv}$ after training qDKT with fastText initialization (left) and DKT with randomized initialization (right). Points are colored red (circle as marker) for correct responses and blue (triangle as marker) for incorrect responses. We observe a clear separation between the correct/incorrect responses for the case of qDKT but see no such separation for DKT. This separation in the embedding space is one factor that leads to the improved predictive performance seen with the fastText initialization scheme.}
\label{fig:reflect1}
\end{figure*}

\begin{figure*}[t!]
\centering
\includegraphics[width=1\textwidth]{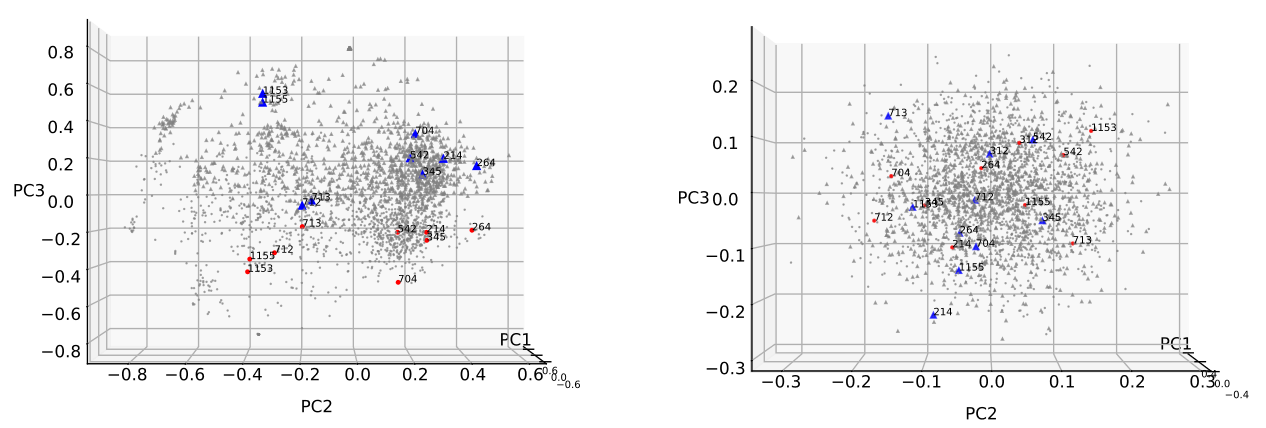}
\caption{Some of the question assessment embeddings have been labeled from Figure~\ref{fig:reflect1}. Question ID are used as labels while color/marker indicates the assessment. In case of fastText initialization (left), one can observe a plane along the third principal component about which the assessment embeddings are approximately reflected.}
\label{fig:reflect2}
\end{figure*}

\section{Empirical Results}
\label{sec:results}

\subsection{Datasets Description}

We consider four datasets for our experiments: ASSISTments 2009, ASSISTments 2017, Statics 2011\footnote{https://pslcdatashop.web.cmu.edu/DatasetInfo?datasetId\\=507}, and a dataset from Tutor -- an online learning platform. The Statics 2011 dataset is from an engineering statics course. Standard pre-processing steps common in the literature are used to clean the data. For ASSISTments2009 dataset, we follow the pre-processing steps recommended by \cite{xiong2016going}. Duplicated records and scaffolding problems are removed. Also, since the dataset contains a few questions that are associated with multiple skills, those multiple skills were combined into a new joint skill for skill-level DKT models, along the lines of \cite{xiong2016going}. However, for qDKT, our Laplacian regularization approach provides needed flexibility when questions fall under multiple skills, doing away with the need of combining multiple skill into one joint skill. For the ASSISTments2017 dataset, all scaffolding problems are filtered out. Relevant statistics for each dataset are given in Table~\ref{tab:my-table}.

\subsection{Experimental Setup and Evaluation Metrics}
Each experiment consists of comparing our proposed qDKT algorithm against the original DKT algorithm for a given dataset. To further quantify the impact of each proposed improvement to the qDKT model we will measure qDKT performance over four different variants: 1) The base qDKT without any regularization and with randomized initialization, 2) qDKT with regularization and randomized initialization, 3) qDKT without regularization but with our proposed initialization scheme and 4) qDKT with both regularization and with our proposed initialization scheme. For all the experiments and datasets, we perform 5-fold cross validation; 70\% data is used for training and the rest for testing. We report the average receiver operating characteristics curve (AUC) score to compare each method. All the models are trained using the Adam optimizer \cite{kingma2014adam} with dropout \cite{srivastava2014dropout} to reduce overfitting.

\subsection{Results and Discussion}
Our results are displayed in Table~\ref{tab:my-table2}. We see that the base qDKT model without regularization and with randomized initialization outperforms the original DKT model on three of the four datasets used. For the ASSISTments 2009 dataset, base qDKT loses by a large margin.  This is due to ASSISTments 2009 dataset having a large number of questions coupled with a low number of observations per question (see Table~\ref{tab:tab1}). 

We note that the individual addition of either the regularizer or the fastText initialization scheme greatly improves the performance of qDKT for each dataset.  We finally note that the combination of both the regularizer and fastText initialization scheme enables qDKT to achieve better performance than DKT for all datasets considered.

\subsection{Visualizations}
For insights into why fastText-inspired question interaction tuple embeddings perform better as compared to those drawn from the standard normal distribution, we performed PCA on the columns of $W_{xv}$ after training qDKT. We then plot the columns of the PCA projection in Figure~\ref{fig:reflect1}. The red points correspond to the question interaction tuple embeddings in which the assessment was correct, while the blue correspond to those in which the assessment was incorrect. In case of fastText-inspired embeddings (Figure~\ref{fig:reflect1}, left), one can observe a clear separation between correct and incorrect responses, whereas no such separation occurs for the case of random initialization.

In Figure~\ref{fig:reflect2}, we label some of the points in Figure~\ref{fig:reflect1} to further illustrate the symmetry we gain through the fastText initialization. The label has the question ID information, while the color/marker indicates the assessment. One can observe that in the case of fastText-inspired initialization (left image), the separating plane (approximately along the third principal component) serves as a reflector for question assessment embeddings.

Symmetry about some plane is expected at the beginning of the training since we initialize the embedding of $(q, 0)$ with the sum of embeddings of $\{f(q), `0\textrm'\, f(q) + `0\textrm'\}$ and likewise for $(q, 1)$ for any question $q$. However, the preservation of the reflecting plane post-training suggests that such a geometry is useful and maintained by the model throughout training.

\section{Conclusion}
\label{sec:conclusions}
We have proposed qDKT, a novel model for knowledge tracing for educational data. Our method improves on prior art by predicting student performance at the question-level, rather than at the skill level.  We have further proposed novel regularization and initialization schemes that greatly improve the performance of our method across several real-world datasets when compared with the traditional knowledge tracing methods. We propose that qDKT can provide a simple, yet tough-to-beat baseline, for new question-centric KT models to come.

\bibliography{bibfile1.bib}
\bibliographystyle{abbrv}
\balancecolumns
\end{document}